\documentclass[conference]{IEEEtran}
\IEEEoverridecommandlockouts
\usepackage{cite}
\usepackage{amsmath,amssymb,amsfonts}
\usepackage{algorithmic}
\usepackage{graphicx}
\usepackage{textcomp}
\usepackage{xcolor}
\usepackage{multirow}
\usepackage[caption = false]{subfig}
\def\BibTeX{{\rm B\kern-.05em{\sc i\kern-.025em b}\kern-.08em
    T\kern-.1667em\lower.7ex\hbox{E}\kern-.125emX}}
\begin{document}

\title{Balancing long- and short-term dynamics for the modeling of saliency in videos\\
\thanks{The authors gratefully acknowledge partial support from the German Research Foundation DFG under project CML~(TRR~169) and Horizon Europe under project TRAIL.}
}

\author{\IEEEauthorblockN{Theodor Wulff\IEEEauthorrefmark{1}, 
Fares Abawi\IEEEauthorrefmark{2}, 
Philipp Allgeuer\IEEEauthorrefmark{2} and
Stefan Wermter\IEEEauthorrefmark{2}}\\
\IEEEauthorblockA{\IEEEauthorrefmark{1}Department of Computer Science \\
The University of Manchester\\
Manchester, United Kingdom \\
theodor.wulff@manchester.ac.uk}\\
\IEEEauthorblockA{\IEEEauthorrefmark{2}Department of Informatics \\
Universität Hamburg\\
Hamburg, Germany \\
\{fares.abawi,philipp.allgeuer,stefan.wermter\}@uni-hamburg.de}

}

\maketitle

\begin{abstract}
The role of long- and short-term dynamics towards salient object detection in videos is under-researched.
We present a Transformer-based approach to learn a joint representation of video\linebreak frames and past saliency information. 
Our model embeds long- and short-term information to detect dynamically shifting saliency in video.
We provide our model with a stream of video frames and past saliency maps, which acts as a prior for the next prediction, and extract spatiotemporal tokens from both modalities.
The decomposition of the frame sequence into tokens lets the model incorporate short-term information from \textit{within} the token, while being able to make long-term connections \textit{between} tokens throughout the sequence.
The core of the system consists of a dual-stream Transformer architecture to process the extracted sequences independently before fusing the two modalities. 
Additionally, 
we apply a saliency-based masking scheme to the input frames to learn an 
embedding that facilitates the recognition of deviations from previous outputs.
We observe that the additional prior information aids in the first detection of the salient location. 
Our findings indicate that the ratio of spatiotemporal long- and short-term features directly impacts the model's performance. 
While increasing the short-term context is beneficial up to a certain threshold, the model's performance greatly benefits from an expansion of the long-term context.

\end{abstract}


\section{Introduction}
Cognitive studies of the human visual system revealed that humans can only process a fraction of the amount of visual stimuli they receive~\cite{frintropComputationalVisualAttention2010}. 
This bottleneck within the human visual pathway requires information filtering to ensure efficient computation.
This and other natural mechanisms have inspired various sub-fields of computer vision, many of which remain active fields of research. 
Predicting and mimicking the human's attention and the corresponding gaze patterns in images has been of great interest to the research fields of psychology, neurobiology and computer science.
The aims of these studies range from understanding human visual attention, and improving the detection and recognition quality of computational vision systems~\cite{frintropComputationalVisualAttention2010} to mimicking human gaze to improve social robot interactions~\cite{fu2024human,allgeuer2024chatty}.

The task of saliency prediction can be defined as the recreation of eye fixations and gaze patterns.
A related task is the extraction of a pixel-level mask of the salient object, generally referred to as salient object detection.
Analyzing saliency in static images is an important task that can lead to improvements in applications in the areas of image segmentation, image cropping, and image classification.
The analysis of saliency in video data aims to deepen the understanding of human attention mechanisms concerning dynamic scenes instead of static images.
It aims to refine applications like video object segmentation and tracking, as well as video classification and retrieval.

The main difference between the two domains is that motion over time which influences the human attention mechanism is not incorporated when detecting saliency in images.
Thus, modeling saliency in videos requires the integration of additional temporal information~\cite{guPyramidConstrainedSelfAttention2020} for a variety of reasons. 
Human attention does not fixate on the same location for long periods of time but shifts between locations based on the given context. 
The effect known as inhibition of return~\cite{kleinIOR} further influences the next fixation by reducing the likelihood of refocusing on the previous location for a concise period.

In frame sequences where the scenery changes over time the shift in human saliency is especially prominent. 
Moreover, relatively small objects moving at high speeds attract the attention of a human observer even though they occupy a minor portion of the image. 
Camera motion can also lead to the object of interest being centered and easily detected in one frame yet occluded in another, resulting in suboptimal predictions.
Approaches are proposed to address saliency shift in videos, and mitigate effects brought about by varied camera and object motion patterns, e.g. by explicitly predicting when the object of interest shifts~\cite{fanShiftingMoreAttention2019}. 
Nevertheless, the mentioned challenges are nowhere near solved. 
It is also unclear which role long- and short-term information play in the modeling of human saliency.
To shed light on this aspect, we study the effects of different temporal context sizes to evaluate their effect on predicting saliency shift.

In this work, we present a Transformer-based model, which investigates the problem of video salient object detection (VSOD).
We hypothesize that the combination of local short-term information and the possibility of embedding long-term dependencies can elevate the performance levels of Transformer-based saliency models.
We extract three-dimensional tokens from the input sequence to capture spatiotemporal short- and long-term information and investigate the relationship between the two. 
To facilitate the creation of subsequent predictions we integrate prior information into the model with the assumption that specifically guiding the model towards salient regions in a video serves as an effective pretraining step.
We provide past saliency information to the model to let it focus on relevant areas in the video to determine whether to keep track of the previous salient object or to shift attention.
The prior information is provided directly in a sequence of past saliency maps, and encoded in the input frames via a saliency-guided input masking scheme.
The latter draws inspiration from successful masked pretraining methods for the training of foundation models~\cite{baevskiData2vecGeneralFramework2022}.
Our main contributions can be summarized in the following way:
\begin{itemize}
    \item \textbf{Temporal context representation.}
    We show that the model benefits most from short-term information if the temporal span remains within a specific range, while as we increase the long-term information, the model's prediction quality continuously improves and has not yet reached a saturation point. 
    \item \textbf{Prior saliency information.}
    We find that providing prior saliency information to the model guides the next prediction towards the past salient region. This relaxes the problem of dynamic video saliency detection and serves as a pretraining step before training the model on downstream tasks.
\end{itemize}

\noindent In Section~\ref{sec:related_work}, we introduce existing work addressing attention modeling in videos or Transformer-based models in vision applications.
In Section~\ref{sec:architecture}, we present the components of our model, while we describe our experimental setup in Section~\ref{sec:experiment}. We investigate and discuss the influences of different temporal contexts on the model performance in Sections~\ref{sec:temporal_context} and~\ref{sec:discussion}.

\section{Related Work}\label{sec:related_work}
\subsubsection{Attention modeling in videos}
Many models encountered in literature on video saliency modeling are built on top of the ConvLSTM layer~\cite{shiConvolutionalLSTMNetwork2015}. Such architectures use a deep CNN block to extract features from the input video and add an attention mechanism to improve network performance.
Fan et al.~\cite{fanShiftingMoreAttention2019} released the Saliency-Shift-Aware VSOD Model as well as a new dataset to specifically address the problem of shifting saliency. The model consists of two main components: a pyramid-dilated convolution module to extract static features and a saliency-shift-aware ConvLSTM module to analyze temporal dynamics and discriminate salient objects from one another.
STA-Net~\cite{biSTANetSpatialtemporalAttention2021} is a spatiotemporal attention (STA) network that utilizes ResNet-50~\cite{heDeepResidualLearning2016} to extract features that are used in 
later stages that generate the saliency maps. 
One stage utilizes multiple attention branches on input slices, while the 
other is a slight modification of the previously mentioned methods of Fan et al.~\cite{fanShiftingMoreAttention2019}.
The PCSA (Pyramid Constrained Self-Attention)~\cite{guPyramidConstrainedSelfAttention2020} network utilizes MobileNetV3~\cite{howardSearchingMobileNetV32019a} as a backbone and implements the restrained self-attention modules in a pyramid architecture. Instead of learning global contexts, PCSA focuses on local motion patterns extracted from a short segment of consecutive frames.
By adding a recursive component to existing Fully Convolutional Networks~\cite{longFullyConvolutionalNetworks2015} 
Wang et al.~\cite{wangSalientObjectDetection2019} take the previous time step as prior into account and are able to elevate the predicted saliency iteratively.
Similarly, the CCNet~\cite{wangCrossComplementaryFusion2020} is based on Fully Convolutional Networks. This mixed architecture consists of a Spatio-Temporal Information Learning Module to capture the cues between frames, while the Single-Image Representation Enhancement module learns static saliency features. By fusing both outputs together, a robust saliency prediction is achieved.
Zhou et al.~\cite{zhou_sti-net_2023} aggregate temporal features from the Optical Flow~\cite{zhouComputationOpticalFlow1988} and combine them with spatial features. They correct the final output by reusing the original input in a final saliency fusion step.
Jian et al.~\cite{jian_flow-edge-net_2023} detect temporal dynamics separately from object boundaries and design a self-adaptive weighted fusion model to integrate the features before the final output.
Zhao et al.~\cite{zhao2024motion} incorporate a space-time memory module to gather the relevant information for salient object detection across adjacent frames.
The work of Huang et al.~\cite{huang2023lightweight} aims to reduce the computational cost of larger VSOD models using a small lightweight model.

Apart from the successes of the previous approaches in the task of VSOD, similar architectures have proven themselves to be successful in Video Saliency Prediction, such as the work of Zhang et al.~\cite{zhangSpatialTemporalRecurrentNeural2021}. They utilize a static and dynamic network branch to extract features before passing them to the ConvLSTM layer.
The amount of frames processed simultaneously is usually limited to the short-term context in a small number of frames. 
We address this issue by investigating the potential of incorporating supplementary long-term information to enhance the model's performance.  

\subsubsection{Transformer architectures for saliency}
After the Transformer architecture proposed in~\cite{vaswaniAttentionAllYou2017} had been successfully applied in language processing, it was later extended to other domains. 
The Vision Transformer by Dosovitskiy et al.~\cite{dosovitskiyImageWorth16x162021} adapts the architecture to process images by splitting each image into a sequence of patches and treating those as input tokens. 
The Video Vision Transformer~\cite{arnabViViTVideoVision2021} extends these patches by a third temporal dimension and makes the model applicable to processing video data.

Liu et al.~\cite{liuVisualSaliencyTransformer2021} develop the Visual Saliency Transformer with respect to salient object detection. 
The image patches that serve as input allow for the extraction of global contexts, while a specific upsampling method and multi-token fusion strategies further aid in the generation of high-resolution results.
Ma et al.~\cite{maVideoSaliencyForecasting2022} extend existing VSP tasks to video saliency forecasting.
Their video saliency forecasting Transformer network
uses the video swin Transformer architecture~\cite{Liu2021VideoST} as an encoder and introduces cross-attention guidance blocks as well as reconstruction blocks in the decoder.
Lou et al.~\cite{lou2022TranSalNet} combine a CNN encoder and Transformer encoder blocks to predict saliency in a given image. 
The CNN encoder is used as a feature extractor, while the Transformer blocks enhance the long-term and contextual information that is passed to the CNN decoder.
Jing et al.~\cite{jing_ebm_sod21} utilize a top-down vision Transformer model and combine it with a learnable energy-based prior in RGB and RGB-D salient object detection tasks.
Zhou et al.~\cite{zhou_Transformer-based_2023} propose a semantic-guided Transformer-encoder and a hierarchical decoder to perform saliency prediction based on spatiotemporal features from different levels.
Su et al.~\cite{su_unified_2024} use the same CNN- and Transformer-based model for Video Salient Object Detection and other group-based segmentation tasks.

All these approaches effectively utilize Transformer architectures for saliency prediction or salient object detection in either images or videos. 
However, examining the interplay between different temporal short- and long-term context representations remains a scarcely emphasized aspect within these studies.

\section{Model Architecture}\label{sec:architecture}
Our model architecture is based on the vision Transformers ViT~\cite{dosovitskiyImageWorth16x162021} and its extension to the video domain ViViT~\cite{arnabViViTVideoVision2021}. 
The choice of a Transformer model has some benefits over convolutional neural networks. 
Transformers are not restrained by the choice of kernel size, which directly impacts the modeling of dependencies between features to gain a global understanding of the input, and are scalable to adhere to different resolutions and lengths of input sequences~\cite{dosovitskiyImageWorth16x162021,vaswaniAttentionAllYou2017}.

\subsection{Video input streams}\label{subsec:tubelet_extraction}
\subsubsection{Input processing} 
The model takes a sequence of current frames and the corresponding past saliency maps as input.
The frames and saliency maps in the input sequences are sorted along the time dimension in ascending order. 
Similar to Arnab et al.~\cite{arnabViViTVideoVision2021}, our model extracts a longer sequence of tubelets across multiple frames from both input sequences. 
A tubelet is the temporal combination of two-dimensional image patches at the same spatial location, extending it to a three-dimensional shape similar to a geometrical cuboid.
Each tubelet contains local short-term spatiotemporal information based on its position and temporal depth. 
The Transformer self-attention layers use the entire sequence of tokens as input, which allows for the detection of inter-tubelet long-term dependencies. 
\begin{figure*}[ht]
    \centering
    \includegraphics[width=.75\textwidth]{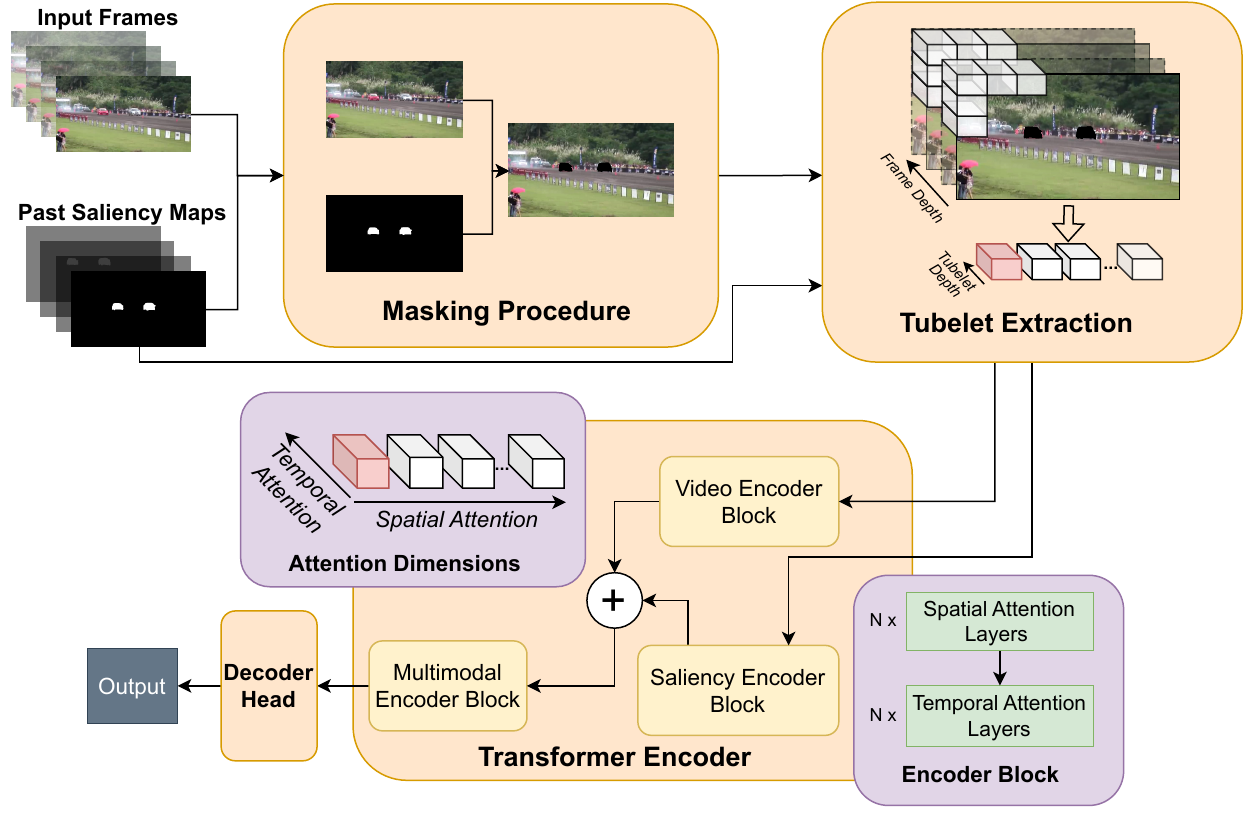}
    \caption{Model components. The past saliency maps are utilized for the masking of the input frames. From both inputs a sequence of tubelets is extracted which serves as input for the domain-specific encoder blocks that attend to the sequence along different dimensions before the interchangeable decoder head generates the final output.}
    \label{fig:model}
\end{figure*}%



\subsubsection{Masking procedure}
Applying masks to the input frames could especially enhance the recognition of a dynamically changing salient region in the current frame.
In order to mask the salient object or region in a frame, we perform a pixel-wise multiplication between each value in the inverted saliency map and the corresponding frame, where each value in the inverted salient object map is either $0$ if part of the salient object or $1$ otherwise. 
By doing so, multiplication keeps the original pixel value if it is not part of the salient object and otherwise sets the pixel values to $0$ or removes a fraction.
During training, the ground-truth saliency maps can be used to mask the video frames. 
In an end-to-end system, the output for frame $t$, the predicted saliency map, is used to then mask the next frame in the sequence $t+1$. 
This extends the input sequence and should teach the network not only to attend to specific areas in the video but also to consider a shift in saliency based on the past frames and saliency maps.
Additionally, the masks can be applied concurrently or with offset $o$. 
The offset encodes the information where attention has been $o$ frames ago in the current frame. 
This emphasizes dynamic changes that become especially apparent between temporally distant frames.

\subsubsection{Tubelet extraction}
We partition the sequence of input frames into tubelets of depth $d_t$ and equal width and height.
The tubelet extraction process begins in the upper left pixel of the first $d_t$ images in the frame sequence. 
Tubelets are then extracted horizontally from left to right within each row.
Once the extraction process has reached the lower right corner it progresses with the upper left corner at the next possible time step. This position is determined by displacing the extraction window $d_t$ frames along the temporal dimension.
The process continues until the last frame has been reached.
 

\subsubsection{Task token}
To allow for differentiating between possible future downstream tasks, we also insert a one-hot encoded task token to the beginning of the extracted tubelet sequence.
Dosovitskiy et al.~\cite{dosovitskiyImageWorth16x162021} utilize a similar token at the beginning of the sequence to make the final classification of the input, since it contains information on the whole sequence after passing through various Transformer layers.
Figure~\ref{fig:model} depicts the tubelet extraction process with the additional task token (red).

\subsection{Transformer model}
\subsubsection{Linear embedding}
The first layer of the encoder maps the input to the specified output dimension by applying a linear transformation to the input along the space dimension of the tubelet sequence.
This results in a mapping of a tubelet of shape $(d_t, c_t, w_t, h_t)$ to $(d_t, dim)$, with $dim$ specifying the output size of the linear layer and $c_t, w_t, h_t$ referring to channels, width, and height of the tubelets respectively.
We chose $dim$ to be 768 to align with the ViT weights and initialize the Transformer layers without the need for additional mapping to reconcile shape discrepancies.
For the video frame and saliency map sequences, two separately trained linear embedding layers are utilized.

\subsubsection{Positional encoding}
Our model utilizes the positional encoding described by Vaswani et al.~\cite{vaswaniAttentionAllYou2017}.
The vector containing the corresponding sinusoids at different positions is pre-computed during the initialization of the model to reduce computational overhead during the forward pass of the model.

\subsubsection{Transformer encoder}
After the tokens have been embedded, the positional encoding is added to those tokens.
The two embedded token sequences are then further processed within their respective Transformer encoder blocks.
The Transformer encoder layers implement the structure of the original Transformer encoder proposed by Vaswani et al.~\cite{vaswaniAttentionAllYou2017} and utilize a configuration that has shown to be effective in the ViT model~\cite{dosovitskiyImageWorth16x162021}.
For the multi-head attention mechanism, the number of heads is set to 12 per default.
The activation function used within each layer is the GELU activation function, and the layer normalization is applied before the internal attention and feed-forward operations.
Before handing the output directly to the decoder head, the tokens of both modalities are summed
and passed into a block of identical Transformer encoder layers, the multimodal encoder that now processes the joint-modality tokens.

\subsubsection{Alternating attention}
The input of shape $(n, dim)$ for the ViT model is a sequence of $n$ embedded tokens.
We apply the embedding along the temporal dimension of the extracted tubelets, resulting in the extended shape $(n, d_t, dim)$.
The authors of the ViViT model~\cite{arnabViViTVideoVision2021} take advantage of the additional dimensionality by computing the attention along the time and space dimensions ($d_t$ and $n$) individually in different steps.
Similarly, our model also processes the input along different dimensions.
The individual Transformer encoder blocks first attend to the space dimension for the first half of the layers within the block and the temporal dimension for the remaining layers.
Since the saliency map encoder, the video frame encoder, and the multimodal encoder are treated as separate blocks, this results in an alternating order of attention computation.
Calculating self-attention~\cite{vaswaniAttentionAllYou2017} along the time dimension is especially relevant to embedding the spatiotemporal local short-term information contained in the individual tubelets. 

\subsubsection{Decoder head}\label{subsec:decoder}
The decoder part of the presented system can be interpreted as interchangeable heads of the architecture.
The specifics of the decoder head depend on the individual configuration.
It is possible to utilize a differently configured decoder head between training on different tasks. 
This lets the model produce flexible outputs that can easily be adapted to suit the current objective.
For the task of pretraining for video salient object detection, we chose a simple convolutional network as the decoder head.
The convolutional neural network decoder head reassembles the token sequence into its original form and passes it through a variable number of convolutional layers, which results in the final output map.

\subsubsection{Frame dropout}
To prevent the model from solely basing predictions on the input stream of saliency maps, we randomly remove some of the input maps during training and replace their value with an all-zero saliency map with probability $p_{drop}$.
This removes the prior at a specific time step and encourages the model to incorporate more features from the video frame stream.
We explored values within $[0.1, 0.8]$ and found a value of $p_{drop}$ of 0.6 to be optimal for our experiments.
Additionally, a binary flag controls whether the frame depth varies during training, with the possible values for the frame depth $d_f$ all being positive multiples $k \cdot d_t \leq d_f$, with $k \in \mathbb{Z}$.
This should let the model put more focus on the most recent frames in the sequence which are always fed into the network.



\section{Experimental Setup}
\label{sec:experiment}
We trained the models 
for a total of 20 epochs each and a batch size of 2, with each sample in the batch containing the sequences of video frames and past saliency maps.  
We set the learning rate to $10^{-5}$ using the Adam optimizer and learning rate cosine annealing with a minimum learning rate of $10^{-7}$ that is reached after $10^{4}$ iterations. 
We resize the input frames to $224 \times 224$ pixels and a width and height of 16 pixels for the tubelets following Dosovitskiy et al.~\cite{dosovitskiyImageWorth16x162021}, while varying the depth.
The model fuses the output of the first six encoder layers for each modality of the input streams by summation before passing it into another six multimodal Transformer encoder layers.  
Since the model returns multiple frames, but only one is actually relevant for a prediction, we calculate the pixel-wise binary cross-entropy on the last frame in the output sequence.
For the decoder, we utilize three convolutional layers with a $3 \times 3$ kernel.
We evaluate the performance of the model on a variety of common metrics to compare salient object maps, including the Mean Average Error (MAE), F-score (with $\beta^2=0.3$), Pearson's Correlation Coefficient (CC) and S Measure~\cite{bylinskiiWhatDifferentEvaluation2019,fanStructuremeasureNewWay2021}.
We train the model on the DAVSOD dataset~\cite{fanShiftingMoreAttention2019} and evaluate it on the three provided test subsets that are split according to the difficulty of the sequences (\textit{Easy}, \textit{Normal} and \textit{Difficult}) as determined by Fan et al.~\cite{fanShiftingMoreAttention2019}.

\section{Results}\label{sec:temporal_context}
Increasing the amount of temporal information accessible to the network could improve the ability to detect a shift in saliency. 
We conducted the following experiments to examine the behaviour of the model towards increases in the number of video frames that serve as global context as well as different local spatiotemporal contexts through tubelet depths.
We provide results for a frame amount of 4, 8 and 12 frames. 
By setting the stride parameter of the frame sequence, we increase the covered frame span by the respective value. 
The chosen stride of 5 leads to the frame sequences spanning 20 (1 second), 40 (2 seconds), and 60 (3 seconds) frames.
Scaling the tubelet size and the amount of input frames with the same factor results in the same amount of deeper tubelets. 
Increasing the tubelet size without changing the amount of input frames shortens the tubelet sequence, and vice-versa. 
Since our model requires previous saliency information as input, we provide previous ground-truth saliency maps to the model to predict the next unknown map and mask the input frames. 
A fine-tuned model could recursively incorporate its past predictions.

\subsection{Increasing the number of input frames}
\label{subsec:increase_input_frames}
At first, the highlighted experiments cover the increase in input frames with a tubelet depth of 2 to investigate whether solely increasing the long-term context also increases the performance.
As such, the results of the model trained on a video sequence with 4, 8 and 12 frames are shown in Table~\ref{tab:res_s4812_d2}.
The model performs best on the normal fraction of the DAVSOD dataset.
The comparative performance to the easy and difficult fractions shows only minor differences for most scores. 
We observe that only the values of the S Measure and MAE of the easy sequences slightly outperform the measurements of the normal sequences.
Overall, a relatively high recall value with comparatively low precision suggests that the model tends to predict larger areas than the actual object to be salient. 
\begin{table*}[ht]
    \centering
    \caption{Evaluation of increased frame depths on the DAVSOD test subsets.}
    \begin{tabular}{l|ccc|ccc|ccc}
    \hline
        $d_f$ & \multicolumn{3}{c|}{4} & \multicolumn{3}{c|}{8} & \multicolumn{3}{c}{12}\\
        $d_t$ & \multicolumn{3}{c|}{2} & \multicolumn{3}{c|}{2} & \multicolumn{3}{c}{2}\\\hline
        Metric    & Easy   & Normal & Difficult & Easy   & Normal & Difficult & Easy   & Normal & Difficult \\\hline
        Precision$\uparrow$ & .335 & .336 & .301    & .330 & \textbf{.337} & .285    & .298 & .318 & .249  \\
        Recall$\uparrow$    & .731 & \textbf{.752} & .685    & .683 & .719 & .680    & .702 & .732 & .716 \\  
        AUC$\uparrow$       & \textbf{.863} & .861 & .837    & .859 & .850 & .832    & .856 & .845 & .827  \\  
        F-score$\uparrow$   & .383  & \textbf{.385}  & .346    & .374 & .384  & .329    & .344 & .365 & .293\\  
        MAE$\downarrow$       & .231 & .253 & .254    & \textbf{.205} & .245 & .254    & .242 & .271 & .302 \\  
        CC$\uparrow$        & .487 & .490 & .445    & \textbf{.500} & .494 & .454    & .497 & .492 & .454 \\  
S Measure$\uparrow$ & \textbf{.520} & .509 & .506    & .516 & .499 & .499   & .489 & .481 & .458 \\\hline
    \end{tabular}
    \label{tab:res_s4812_d2}
\end{table*}
After increasing the frame depth, the overall measured performance drops in almost all reported metrics except the MAE on the easy and normal sequences and the CC.
The decreasing trend observed between the previous two configurations continues as the number of input frames is increased to 12.
The CC still exceeds the reported value for the configuration with four input frames but drops in comparison to the previous configuration with a sequence length of 8.

\begin{table*}[ht]
    \centering
    \caption{Evaluation of increased tubelet depths on the DAVSOD test subsets.}
    \begin{tabular}{l|ccc|ccc|ccc}
    \hline
        $d_f$ & \multicolumn{3}{c|}{4} & \multicolumn{3}{c|}{8} & \multicolumn{3}{c}{8}\\
        $d_t$ & \multicolumn{3}{c|}{4} & \multicolumn{3}{c|}{4} & \multicolumn{3}{c}{8}\\\hline
        Metric    & Easy   & Normal & Difficult & Easy   & Normal & Difficult & Easy   & Normal & Difficult\\\hline
        Precision$\uparrow$ & \textbf{.489} & .484 & .432     & .450 & .461 & .421     & .380 & .381 & .321\\
        Recall$\uparrow$    & .706 & .755 & .698    & .619 & .675 & .604    & .700 & \textbf{.757} & .755 \\  
        AUC$\uparrow$       & .902 & .904 & \textbf{.906}   & .869 & .869 & .861    & .876  & .869 & .878 \\   
        F-score$\uparrow$   & .526 & \textbf{.528} & .473    & .481 & .498 & .453    & .425  & .430  & .370 \\  
        MAE$\downarrow$       & \textbf{.098} & .134 & .121    & .115 & .153 & .138    & .152  & .206  & .200 \\  
        CC$\uparrow$        & \textbf{.525} & .516 & .482    & .474  & .468 & .443    & .412 & .372 & .403 \\  
S Measure$\uparrow$ & \textbf{.572} & .561 & .536   & .556 & .551 & .526    & .531 & .519 & .498 \\\hline
    \end{tabular}
    \label{tab:res_s8_d48}
\end{table*}
\subsection{Increasing the tubelet depth}
\begin{figure*}[!ht]
\captionsetup[subfigure]{labelformat=empty}
    \centering
    \subfloat{\includegraphics[width=0.35\textwidth]{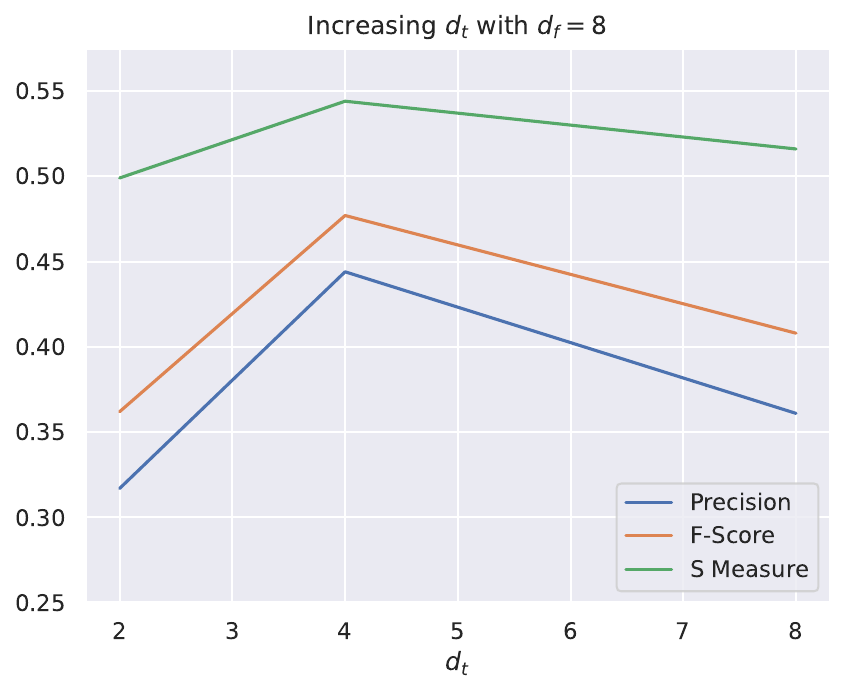}}
    \subfloat{\includegraphics[width=0.35\textwidth]{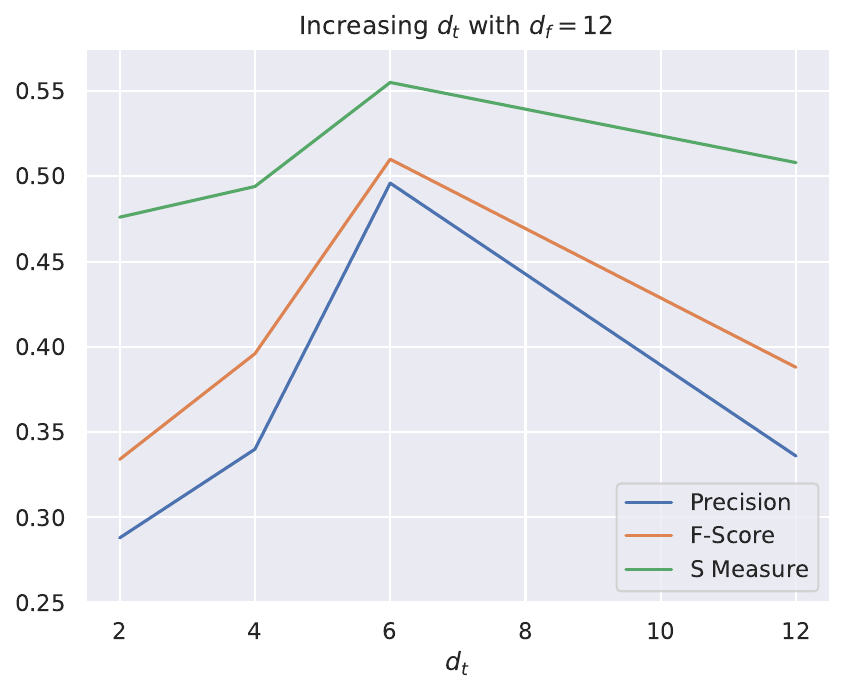}}
    \caption{Precision, F-Score and S Measure across different $d_t$ for $d_f=8$ and $d_f=12$.}
    \label{fig:metric_plots}
\end{figure*}

\begin{figure*}
\captionsetup[subfigure]{labelformat=empty}
    \centering
    \subfloat{\includegraphics[width=0.11\textwidth]{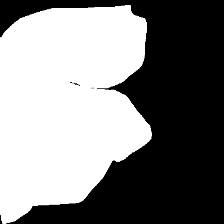}}\hspace{0.05em}
    \subfloat{\includegraphics[width=0.11\textwidth]{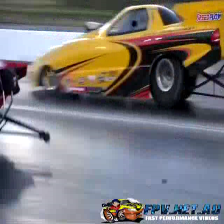}}\hspace{0.05em}
    \subfloat{\includegraphics[width=0.11\textwidth]{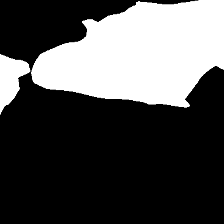}}\hspace{0.05em}
    \subfloat{\includegraphics[width=0.11\textwidth]{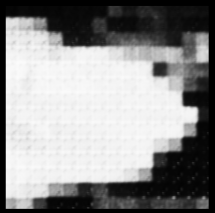}}\hspace{0.05em}
    \subfloat{\includegraphics[width=0.11\textwidth]{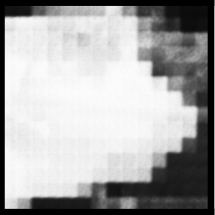}}\hspace{0.05em}
    \subfloat{\includegraphics[width=0.11\textwidth]{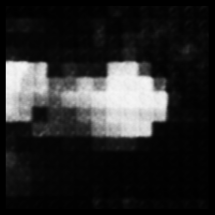}}\hspace{0.05em}
    \subfloat{\includegraphics[width=0.11\textwidth]{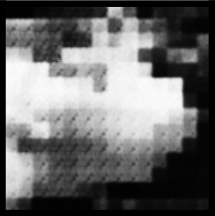}}\\[-0.05em]

    \subfloat{\includegraphics[width=0.11\textwidth]{img/samples/0201_gt.png}}\hspace{0.05em}
    \subfloat{\includegraphics[width=0.11\textwidth]{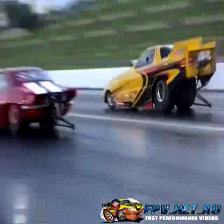}}\hspace{0.05em}
    \subfloat{\includegraphics[width=0.11\textwidth]{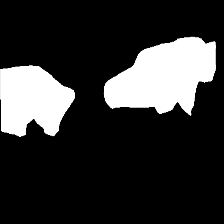}}\hspace{0.05em}
    \subfloat{\includegraphics[width=0.11\textwidth]{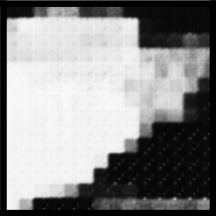}}\hspace{0.05em}
    \subfloat{\includegraphics[width=0.11\textwidth]{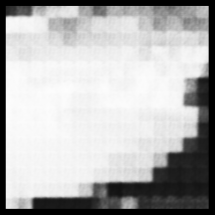}}\hspace{0.05em}
    \subfloat{\includegraphics[width=0.11\textwidth]{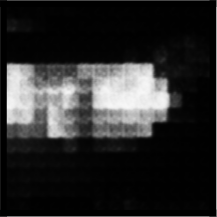}}\hspace{0.05em}
    \subfloat{\includegraphics[width=0.11\textwidth]{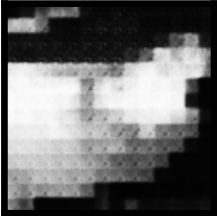}}\\[-0.05em]

    \subfloat{\includegraphics[width=0.11\textwidth]{img/samples/0206_gt.png}}\hspace{0.05em}
    \subfloat{\includegraphics[width=0.11\textwidth]{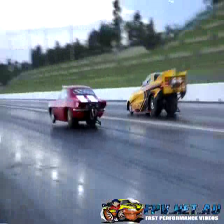}}\hspace{0.05em}
    \subfloat{\includegraphics[width=0.11\textwidth]{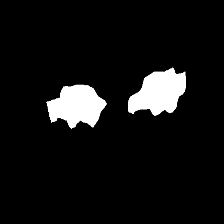}}\hspace{0.05em}
    \subfloat{\includegraphics[width=0.11\textwidth]{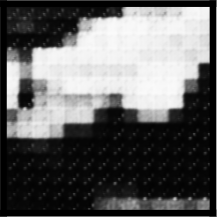}}\hspace{0.05em}
    \subfloat{\includegraphics[width=0.11\textwidth]{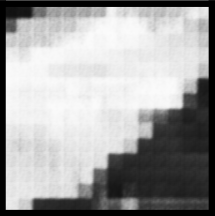}}\hspace{0.05em}
    \subfloat{\includegraphics[width=0.11\textwidth]{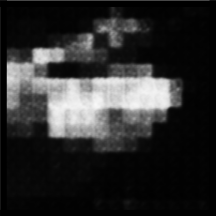}}\hspace{0.05em}
    \subfloat{\includegraphics[width=0.11\textwidth]{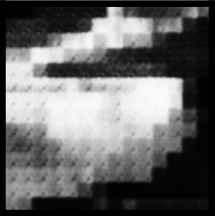}}\\[-0.05em]

    \subfloat[\scriptsize{Context map}]{\includegraphics[width=0.11\textwidth]{img/samples/0211_gt.png}}\hspace{0.05em}
    \subfloat[\scriptsize{Final frame}]{\includegraphics[width=0.11\textwidth]{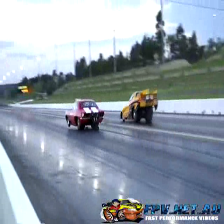}}\hspace{0.05em}
    \subfloat[\scriptsize{Target Map}]{\includegraphics[width=0.11\textwidth]{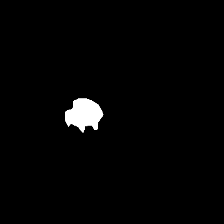}}\hspace{0.05em}
    \subfloat[\scriptsize{$d_t=2$}]{\includegraphics[width=0.11\textwidth]{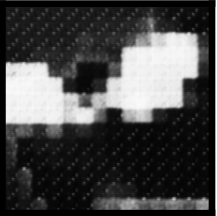}}\hspace{0.05em}
    \subfloat[\scriptsize{$d_t=4$}]{\includegraphics[width=0.11\textwidth]{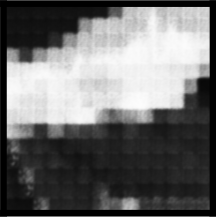}}\hspace{0.05em}
    \subfloat[\scriptsize{$d_t=6$}]{\includegraphics[width=0.11\textwidth]{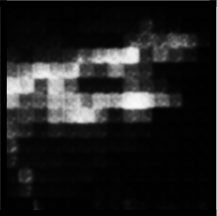}}\hspace{0.05em}
    \subfloat[\scriptsize{$d_t=12$}]{\includegraphics[width=0.11\textwidth]{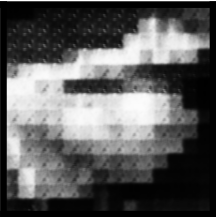}}
    \caption{Output of different tubelet depths with an input sequence of 12 frames.}
   \label{fig:seqlen_output}
\end{figure*}

Increasing the tubelet depth leads to fewer tubelets with more temporal context, while a decrease leads to more tubelets with less context.
When alternating between the temporal and spatial dimensions within the encoder of the network, the layers that compute attention along the time dimension gain more temporal context, while the layers attending to the spatial dimension receive fewer tubelets.
This results in a trade-off between frame depth and tubelet depth.
The subsequent effects are illustrated in Tables~\ref{tab:res_s8_d48} and~\ref{tab:res_s12_d46} when the frame depth is 8 and tubelet depths are 4 and 8, or the frame depth of 12 and tubelet depths of 4, 6 and 12.
Increasing the tubelet depth to 12 with a sequence length of 6 results in the overall highest precision.
It can be seen that the notion of some metrics being higher on the sequences of normal difficulty than on the easy sequences remains prevalent here. 
\begin{table*}[ht]
    \centering
    \caption{Evaluation of increased tubelet depths on the DAVSOD test subsets.}
    \begin{tabular}{l|ccc|ccc|ccc}
    \hline
        $d_f$ & \multicolumn{3}{c|}{12}  & \multicolumn{3}{c|}{12} & \multicolumn{3}{c}{12}\\
        $d_t$ & \multicolumn{3}{c|}{4}  & \multicolumn{3}{c|}{6} & \multicolumn{3}{c}{12}\\\hline
        Metric    & Easy  & Normal & Difficult & Easy   & Normal & Difficult & Easy   & Normal & Difficult\\\hline
        Precision$\uparrow$ & .355 & .354 & .312    & .503 & \textbf{.516} & .469   & .357 & .354 & .298 \\
        Recall$\uparrow$    & .840 & \textbf{.896} & .852    & .549 & .604 & .535   & .779 & .830 & .833 \\  
        AUC$\uparrow$       & \textbf{.919} & .918 & .911    & .872 & .877 & .873   & .885 & .890 & .898 \\   
        F-score$\uparrow$   & .410 & .412 & .365    & .513 & \textbf{.534} & .483   & .408 & .408 & .349  \\  
        MAE$\downarrow$       & .232 & .284 & .289    & \textbf{.096} & .128  & .112   & .165 & .215  & .212 \\  
        CC$\uparrow$        & \textbf{.571} & .565 & .531    & .465 & .461 & .431   & .477 & .452 & .454 \\  
S Measure$\uparrow$ & .509 & .496 & .476    & .566  & \textbf{.564} & .536   & .524 & .511 & .490 \\\hline
    \end{tabular}
    \label{tab:res_s12_d46}
\end{table*}%
The performance decreases again once the tubelet depth matches the frame depth. 
These particular configurations of the system lead to the spatial dimension of the extracted tubelet sequence spanning exactly one time step. 
The configuration with $d_f=4$ is the only exception to this observation. 
Here, the temporal context of $d_t=2$ might be insufficient so that the increase to 4 still improves the performance. 
The collective results of the configurations presented in this set of experiments are summarized in Table~\ref{tab:res_slen_sum}. 
\begin{table*}[!ht]
    \centering
    \caption{Evaluation on the DAVSOD test subsets.}
    \begin{tabular}{l|cc|ccc|cccc}    
    \hline
        $d_f$  & 4     & 4      & 8     & 8     & 8     & 12    & 12    & 12    & 12 \\
        $d_t$ & 2     & 4      & 2     & 4     & 8     & 2     & 4     & 6     & 12 \\\hline
        Precision$\uparrow$     & .324 & .468  & .317 & .444 & .361 & .288 & .340 & \textbf{.496} & .336 \\
        Recall$\uparrow$        & .722 & .720  & .694 & .633 & .737 & .717 & \textbf{.863} & .563 & .814 \\  
        AUC$\uparrow$           & .853 & .904   & .847 & .866 & .874 & .843 & \textbf{.916}  & .874 & .891 \\  
        F-score$\uparrow$       & .371 & .509  & .362 & .477 & .408 & .334 & .396 & \textbf{.510} & .388 \\  
        MAE$\downarrow$           & .246 & .118  & .235 & .135 & .186 & .271 & .268 & \textbf{.112} & .198 \\  
        CC$\uparrow$            & .474 & .508  & .483 & .462 & .395 & .481 & \textbf{.556} & .452 & .461 \\  
    S Measure$\uparrow$ & .506 & \textbf{.556}  & .499 & .544 & .516 & .476 & .494 & .555 & .508 \\\hline
    \end{tabular}
    \label{tab:res_slen_sum}
\end{table*}
We observe that adding local short-term information by increasing the tubelet depth has a limit, which is reached for most configurations around $d_t = 6$.
This peak is depicted in Figure~\ref{fig:metric_plots}. 
The presented configurations also suggest that incrementing the frame depth allows for deeper tubelets before the observed performance drop occurs.
Since the sequence after the tubelet extraction step already contains a relatively high amount (512 for $d_f=12, d_t=6$) of tubelets, we argue that the task to connect elements in this sequence along the spatial dimension is more complex than along the temporal dimension ($d_t$).
Doubling the temporal dimension does not increase the absolute task complexity drastically compared to doubling the spatial dimension, but it provides more relevant information to the corresponding layers.
Taking the trade-off between the two dimensions into account, it is possible to further increase the model performance by enlarging the temporal context. 
A tubelet with depth $>$ 2 performed generally better in comparison to configurations with shallower tubelets. 
This leads us to the assumption that 2 consecutive frames do not contain sufficient information to adequately extract the video dynamics.
Even with the added stride, 2 frames would span less than 0.5 seconds of video. 
The effect is amplified by the observation that gaze tends to shift at different viewing durations (0.5, 3 and 5 seconds) even when the scene does not change~\cite{foscoHowMuchTime}.
Figure~\ref{fig:seqlen_output} provides examples of the effect of different tubelet depths on a sequence with 12 frames and highlights the trade-off between frame and tubelet depth. 
The model generally shows signs of predicting the protruding edges surrounding the saliency prior as salient. 
Smaller tubelet depths tend to show little detail on the extended prior regions, while deeper tubelets seem to carry more information that allows for a fine-grained prediction of the salient region close to the previously salient locations.  
The tubelet extraction process along hard borders also leads to artifacts in the final output.
    
\section{Discussion}\label{sec:discussion}
The Transformer encoder architecture used within our work is able to process sequences whose length is only limited by the available memory, which can be beneficial when modeling long-term connections in a continuous video stream. 
Without increasing the tubelet depth, the additional input frames increase the tubelet amount after the tubelet extraction step, which only affects the encoder blocks that compute the attention along the spatial dimension of the input.
Additionally, the dynamics enclosed in a single tubelet seem to carry more information content once a specific time span is covered by its depth, mitigating the need for adaptation of the tubelet depth. 
We found that the model makes its predictions in many cases based on the introduced prior saliency information.
It tends to extend the given prior region towards the new salient object's location,
but is not able to accurately model the inhibition of return~\cite{kleinIOR}. 
Nevertheless, this also leads to an enhanced capability of tracking a moving salient object through the knowledge of its past locations (see Figure~\ref{fig:seqlen_output}, $d_t=6$).

\section{Conclusion}
We proposed a novel multimodal Transformer model for saliency prediction tasks and evaluated the effects of extracting different temporal short- and long-term context representations.
We included prior saliency information to guide the learning process towards the salient regions and embed past information to guide the current prediction.
In our assessment, the complexity of modeling relationships along the shorter temporal dimension is lower than along the long spatial dimension.
This led to us uncovering a relationship between the balance of the long- and short-term contexts. 
The optimal performance in our experiments was achieved by providing the model with short-term spatiotemporal inputs that spanned approximately half the depth of our selected long-term temporal range.

\bibliographystyle{IEEEtran}
\bibliography{references}
\end{document}